\pdfoutput=1

\documentclass{llncs}
\usepackage[T1]{fontenc}    
\usepackage{lmodern}
\usepackage{hyperref}       
\usepackage{url}            
\usepackage{booktabs}       
\usepackage{amsfonts}       
\usepackage{amsmath}
\usepackage{nicefrac}       
\usepackage{microtype}      
\usepackage{xcolor}         
\usepackage{xspace}
\usepackage{subcaption}
\usepackage{comment}
\usepackage{tikz}
\usepackage{cleveref}
\usepackage{multirow}

\hypersetup{colorlinks=true,
            urlcolor=blue,
            linkcolor=black,
            citecolor=blue,
            breaklinks=true}

\title{Recent Advances of Differential Privacy in Centralized Deep Learning: A Systematic Survey}

\author{%
  Lea Demelius\inst{1,2}\and
  Roman Kern\inst{1,2}\and
  Andreas Tr\"ugler\inst{1,2,3}
}%

\institute{%
  Graz University of Technology (Austria)\and
  Know-Center GmbH (Austria)\and
  University of Graz (Austria)\newline
  ldemelius@know-center.at\newline
  rkern@know-center.at\newline
  atruegler@know-center.at
}%

\begin{document}

\maketitle

\begin{abstract}
 Differential Privacy has become a widely popular method for data protection in machine learning, especially since it allows formulating strict mathematical privacy guarantees. This survey provides an overview of the state-of-the-art of differentially private centralized deep learning, thorough analyses of recent advances and open problems, as well as a discussion of potential future developments in the field. Based on a systematic literature review, the following topics are addressed: auditing and evaluation methods for private models, improvements of privacy-utility trade-offs, protection against a broad range of threats and attacks, differentially private generative models, and emerging application domains.
\end{abstract}

\section{Introduction}
Deep learning is the state-of-the-art of machine learning used to solve complex tasks in various fields, including computer vision and natural language processing, as well as applications in different domains ranging from healthcare to finance. As the performance of these models relies on a high amount of training data, the rise of deep learning comes with an increased interest in collecting and analysing more and more data. However, data often includes personal or confidential information, making privacy a pressing concern. This development is also reflected in legislation that was recently put in place to protect personal information and avoid identification of individuals, e.g., the General Data Protection Regulation (GDPR) in the European Union or the California Consumer Privacy Act (CCPA).

In response, privacy-enhancing technologies are growing in popularity. Cryptographic techniques \cite{rechberger_privacy-preserving_2022} like homomorphic encryption and secure multi-party computation are used to protect against direct information leakage during data analysis. However, sensitive information can still be leaked indirectly via the output of the analysis and compromise privacy. For example, a machine learning model trained to diagnose a disease might make it possible to reconstruct specific information about individuals in the training dataset, even when the computation is encrypted. To avoid this kind of privacy leaks, output privacy-preserving techniques can be applied. One common approach is to use anonymization techniques like k-anonymity \cite{sweeney_k-anonymity_2002}, l-diversity \cite{machanavajjhala_l-diversity_2007} or t-closeness \cite{li_t-closeness_2007}. However, it is now well known that anonymization often cannot prevent re-identification \cite{backstrom_wherefore_2007,ganta_composition_2008,narayanan_robust_2008} and the privacy risk is not quantifiable. Hence, differential privacy (DP) \cite{dwork_algorithmic_2014} was proposed to provide output privacy guarantees. DP is a mathematical probabilistic definition that makes it possible to hide information about each single datapoint (e.g., about each individual) while allowing inquiries about the whole dataset (e.g., a population) by adding curated noise. While DP provides a good utility-privacy trade-off in many cases (especially statistical queries and traditional machine learning methods like logistic regression or support vector machines), the combination of differential privacy and deep learning poses many challenges arising from the high-dimensionality, the high number of training steps, and the non-convex objective functions in deep learning. As a response, considerable progress has been made in the past few years addressing the difficulties and opportunities of differentially private deep learning (DP-DL).

\subsection{Contributions of this survey}
This survey provides a comprehensive analysis of recent advances in differentially private deep learning (DP-DL), focusing on centralized deep learning. Distributed, federated and collaborative deep learning methods and applications have their unique characteristics and challenges, and deserve a separate survey. We also specifically focus on methods that do not presume convex objective functions.

Our main contributions are:
\begin{enumerate}
    \item the thorough systematic literature review of DP-DL
    \item the identification of the research focuses of the last years (2019-2023): 1) evaluating and auditing DP-DL models, 2) improving the privacy-utility trade-off of DP-DL, 3) DP-DL against threats other than membership and attribute inference, 4) differentially private generative models, and 5) DP-DL for specific applications
    \item the analysis and contextualization of the different research trends including their potential future development
\end{enumerate}

Previous reviews of deep learning and differential privacy (see Table \ref{tab:surveys}) primarily cover the privacy threats specific to deep learning and the most common differential privacy methods used for protection. In contrast, this survey goes beyond the basic methods and systematically investigates advances and new paths explored in the field since 2019. We provide the reader with advanced understanding of the state-of-the-art methods, the challenges, and opportunities of differentially private deep learning.

\begin{table}
\centering
\caption{Reviews on differentially private deep learning (DL-DP).}
\label{tab:surveys}
\makebox[\textwidth]{\begin{tabular}{p{4.5cm} lc p{7.2cm}}
\toprule
\textbf{Review}	& \textbf{Year   } & \textbf{Systematic   } & \textbf{Contribution} \\
\midrule
Zhao et al. \cite{zhao_differential_2019} & 2019 & - &  privacy attacks on DL, basic DP-DL methods\\
Ha et al. \cite{ha_differential_2019} & 2019 & - & privacy attacks on DL, basic DP-DL methods\\
Shen and Zhong \cite{shen_analysis_2021} & 2021 & - & comparison of basic DP-DL models\\
Ouadrhiri and Abdelhadi \cite{ouadrhiri_differential_2022}  & 2022 & - &  DP variants and mechanisms for deep and federated learning\\
\textbf{This survey} & 2023 & \checkmark & recent advances in DP-DL: evaluation and auditing, privacy-utility trade-off, threats and attacks, synthetic data and generative models, open problems\\
\bottomrule
\end{tabular}}
\end{table}

Additional to the reviews mentioned above, a range of broader reviews exist, e.g., about differential private machine learning (without focusing on deep learning) \cite{gong_survey_2020,zhu_more_2022,blanco-justicia_critical_2023}, privacy-preserving deep learning (without focusing on differential privacy) \cite{chang_privacy_2018,tanuwidjaja_survey_2019,ha_security_2020,mireshghallah_privacy_2020,boulemtafes_review_2020,vasa_deep_2022}, or privacy-preserving machine learning in general \cite{liu_when_2021,xue_machine_2020}. These papers give a good overview, while our survey follows a more detailed approach focusing on recent developments of differential privacy in centralized deep learning.

\section{Survey methodology}
\label{sec:methodology}
\begin{figure}[h]
  \centering
  \makebox[\textwidth]{\includegraphics[width=1.4\linewidth]{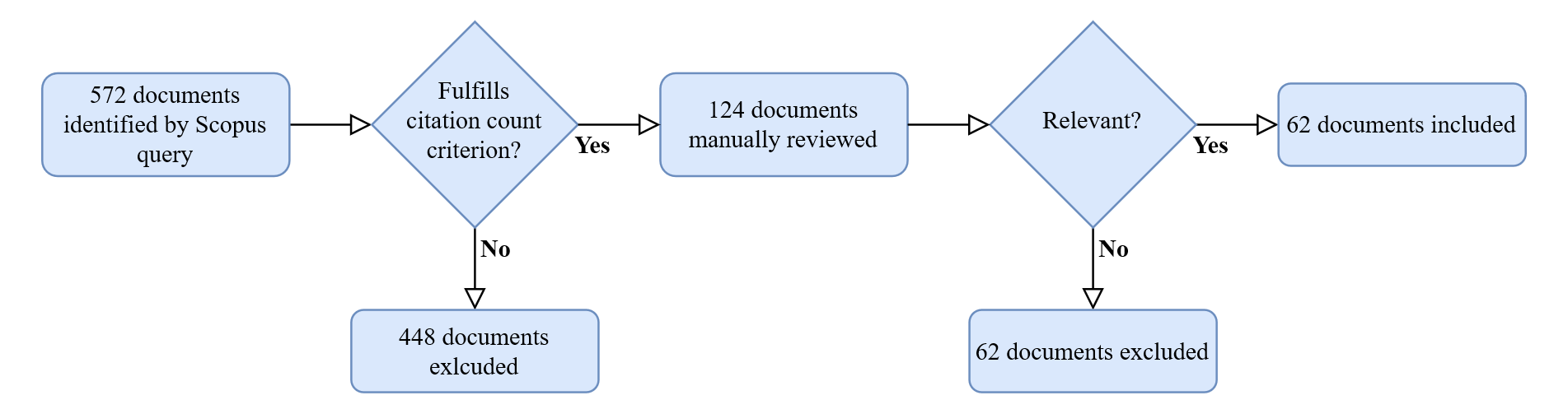}}
  \caption{Flow diagram of the paper selection process.}
    \label{fig:systematic_review}
\end{figure}

 We conducted a systematic literature search on Scopus using a predefined search query, and subsequent automatic and manual assessment according to specific inclusion/exclusion criteria. The full process is depicted in Figure \ref{fig:systematic_review}.
 
 The Scopus query filtered for documents with the keywords "differential privacy," "differentially private," or "differential private" combined with "neural network," "deep learning," or "machine learning," but without mention of "federated learning," "collaborative," "distributed," or "edge" in their title or abstract. The documents were restricted to articles, conference papers, reviews, or book chapters published in journals, proceedings, or books in computer science or engineering. Only documents published in English between 2019 and March 2023 were considered.

Additionally, documents were only included if they were cited at least ten times or were in the top 10\% of most cited papers of the corresponding year. This inclusion criterion allowed manual review of the most influential works in the field. The second part of the criterion was added to mitigate the bias towards older papers.

In the last step, the remaining documents were manually reviewed. Whether a document was included in this survey was decided based on the following criteria: First, differential privacy must be a key topic of the study. Second, the methods must include neural networks with results relevant for deep learning. Third, we only included works about centralized learning in contrast to distributed learning. Fourth, we excluded reinforcement learning and focused on the supervised and unsupervised learning paradigms typical for centralized deep learning. Last, one document was excluded because of retraction.

\section{Preliminaries}

\subsection{Differential Privacy (DP)}
\label{sec:DP}
Differential Privacy (DP) \cite{dwork_algorithmic_2014} is a mathematical definition of privacy that aims at protecting details about individuals while still allowing general learnings from their data. A randomized algorithm $\mathcal{M}:\mathcal{D}\rightarrow \mathcal{R}$ with domain $\mathcal{D}$ and range $\mathcal{R}$ is $\epsilon$-differentially private if for any two datasets $x, y \in \mathcal{D}$ differing on at most one datapoint, and any subset of outputs $\mathcal{S} \subseteq \mathcal{R}$, it holds that

\begin{equation}
\label{eq:DP}
    Pr[\mathcal{M}(x)\in \mathcal{S}]\leq e^{\epsilon}Pr[\mathcal{M}(y)\in \mathcal{S}]
\end{equation}

where $Pr[]$ denotes the probability, and $\epsilon$ is the privacy risk (also referred to as the privacy budget). Simply put, a single datapoint (usually corresponding to an individual) has only a limited impact on the output of the analysis. $\epsilon$ quantifies the maximum amount of information the output of the algorithm can disclose about the datapoint. Therefore, a lower $\epsilon$ results in stronger privacy.

DP is achieved by computing the sensitivity of the algorithm in question to the absence/presence of a datapoint and adding random noise accordingly. The most common noise distributions used for DP are Laplace, Gaussian, and Exponential distribution. The amount of added noise relative to the sensitivity determines the privacy risk.

Differential privacy is not only quantifiable but also has a number of other benefits: For one, it is independent of both the input data and auxiliary information such as publicly available data or future knowledge. Moreover, it is immune to post-processing, and composable, i.e., the sequential or parallel application of two $\epsilon$-differentially private algorithms is at least $2\epsilon$-DP. Advanced composition theorems can even prove lower overall privacy bounds. Another advantage of DP is its flexibility: The noise can be applied at different points in the data flow, for example on the input data, on the output of the algorithm, or somewhere in between (e.g., during training of a deep learning model). It can also be combined with other privacy-enhancing technologies like homomorphic encryption, secure multi-party computation, or federated learning.

While these advantages lead to the widespread acceptance of differential privacy as the gold standard for privacy protection, it also comes with its challenges. Firstly, the unitless and probabilistic privacy parameter $\epsilon$ is difficult to interpret, and thus choosing an appropriate value is challenging. The interested reader is referred to \cite{desfontaines_list_2021} for a list of values used in real-world applications.

Another challenge of DP is its influence on the algorithm’s outputs and, consequently, its properties. The dominant issue is the privacy-utility trade-off. Nevertheless, it can also influence the performance, usability, fairness, or robustness of the algorithm.

Additionally, DP is not easy to understand for laypeople: 1) It is important to note that DP does not offer perfect privacy. DP provides a specific interpretation of privacy, but depending on the context, other interpretations might be expected or more relevant. Besides, the privacy protection depends both on the privacy parameter $\epsilon$ and the unit/granularity of privacy, i.e., on what is considered as one record. That is to say, merely stating that an algorithm is DP does not ensure a substantial guarantee. 2) As DP is a mathematical definition that only defines the constraints but does not mandate an implementation, there exist many different algorithms that satisfy DP, from simple statistical analyses to machine learning algorithms. Some assume a central trusted party that has access to all the data (global DP), others apply noise before data aggregation (local DP). 3) There exist a wide range of DP variants and extensions. Just within the years 2008 and 2021, over 250 new notions were introduced \cite{desfontaines_sok_2022}. They differ, for example, in how they quantify privacy loss, which properties are protected, and what capabilities the attacker is assumed to have. The most common extension is approximate DP (often simply referred to as $(\epsilon,\delta)$-DP), where the algorithm is $\epsilon$-DP with a probability of $1-\delta$. The failure probability $\delta$ is typically set to less than the inverse of the dataset size. Other common relaxations include zero-concentrated DP (zCDP) \cite{dwork_concentrated_2016} and Rényi DP \cite{mironov_renyi_2017}.

\subsection{Deep neural networks and stochastic gradient descent (SGD)}
Deep neural networks consist of connected nodes organised in layers, namely an input layer, multiple hidden layers, and an output layer. Each node applies a set of weights to its inputs and passes its sum through a non-linear activation function. The network can learn to perform different tasks, e.g., classification on complex data by minimizing the loss (i.e., the difference between the predictions of the model and the desired outputs). As the objective function of deep neural networks are often non-convex, it is generally not feasible to solve this optimization problem analytically. Instead, iterative optimization algorithms are applied, most commonly variants of stochastic gradient descent (SGD): At each time step $j$, the model weights $w$ are updated according to the gradients of the objective function $\mathcal{L}$ computed on a randomly selected subset of the training dataset (=batch) $\mathcal{B}$. In Equation \ref{eq:SGD}, $x_i$ denotes a record from the training set with the corresponding target output $y_i$, $\eta$ is the learning rate, and $|\mathcal{B}|$ the batch size. 

\begin{equation}
\label{eq:SGD}
w_j = w_{j-1} - \eta \frac{1}{|\mathcal{B}|} \sum_{i\in \mathcal{B}} \nabla\mathcal{L}(w_{j-1}, x_i, y_i)
\end{equation}

After training the model, e.g., with SGD, neural networks can be used to make predictions for previously unseen data. Models that perform well on new data are well-generalized, while models that perform well only on the training data are considered overfitted.

Deep learning models are usually trained with a supervised or an unsupervised learning paradigm. In supervised learning, the training set is labeled, while in unsupervised learning the model learns to identify patterns without access to a ground truth. Apart from the fully-connected layers described above, neural networks often include other types of architectures and layers. One popular example are convolutional layers, which use filters to extract local patterns, for example edges in images. 

\subsection{Privacy threats for deep learning models}
Privacy attacks on deep learning models target either the training data or the model itself. They can take place in the training or the inference phase. During training, the adversary can not only be a passive observer but also actively change the training process. Moreover, the attacker can have access to the the whole model, i.e., its architecture, parameters, gradients and outputs (white-box access), or only the model's outputs (black-box access). Typical privacy attacks include:

\textbf{Membership inference attack (MIA):} The adversary attempts to infer whether a specific record was part of the training dataset. This type of attack exists both in the white- and black-box setting, where in the latter case one distinguishes further between access to the prediction vector and access to the label. The first (and still widely used) MIA on machine learning models was proposed by Shokri et al. \cite{shokri_membership_2017}. They train shadow models that imitate the target model and based on their outputs on training and non-training data, a classification model learns to identify which records are members of the training data. The attack performs best on overfitted models \cite{yeom_privacy_2018,shokri_membership_2017}.

\textbf{Model inversion attack:} The adversary tries to infer sensitive information about the training data, for example an attribute of a specific datapoint or features that characterize a certain class.

\textbf{Property inference attack:} The attacker seeks to extract information about the training data that is not related to the training task. For example, they might be interested in statistical properties like the ratio of training samples that have a certain property.

\textbf{Model extraction attack:} The adversary learns a model that approximates the target model. While this threat only targets the model and not the training data, it can still increase the privacy risk as a successful model extraction can facilitate follow-up attacks like model inversion.

In is important to note that especially in the case of model inversion and property inference the terms are not used consistently. For example, De Cristofaro \cite{de_cristofaro_critical_2021} uses the term property inference for inferring features of a class, while others \cite{rigaki_survey_2021,park_attack-based_2019,zhang_secret_2020} (including us) refer to this as a type of model inversion.\\

Additionally, there are attacks not specific to privacy that could still be a threat to deep learning models:

\textbf{Adversarial attack:} During the inference phase, the attacker manipulates inputs purposefully so that the model misclassifies them.

\textbf{Poisoning attack:} The adversary perturbs the training samples so that they manipulate the model with the goal to reduce its accuracy or trigger specific misclassifications.\\

For a more detailed description, we refer the reader to surveys that focus on privacy attacks \cite{de_cristofaro_critical_2021,rigaki_survey_2021,shafee_privacy_2021}. A selection of attack implementations can, e.g., be found in the \textit{Adversarial Robustness Toolbox} \cite{art2018}.

\subsection{DP algorithms for deep learning}
\label{sec:DPDL}

The most popular algorithm for DP-DL is differentially private stochastic gradient descent (DP-SGD) by Abadi et al. \cite{abadi_deep_2016}. It adapts classical SGD by perturbing the gradients with Gaussian noise. As the gradients norms are unbounded, they are clipped before to ensure a finite sensitivity. Equation \ref{eq:SGD} thus becomes Equation \ref{eq:DPSGD}, where $clip[]$ denotes the clipping function that clips the per-example gradients so that their norm does not exceed the clipping norm $C$, $\chi$ is a random vector drawn from a standard Gaussian distribution, and $\sigma$ is the standard deviation of the added Gaussian noise.\\

\begin{equation}
\label{eq:DPSGD}
w_j = w_{j-1} - \eta \frac{1}{|\mathcal{B}|} \sum_{i\in \mathcal{B}} clip\left[\nabla\mathcal{L}(w_{j-1}, x_i, y_i), C\right] + \frac{\sigma C}{|\mathcal{B}|}\chi
\end{equation}

While the privacy-utility trade-off inherent to DP still applies (i.e., information is lost), DP-SGD can improve generalization and thus improve accuracy on the validation dataset.

It is important to note that the per-example gradient clipping in DP-SGD used to bind each record's sensitivity differs from the the batch-wise gradient clipping sometimes used to improve stability and convergence of the training process \cite{zhang_why_2020}.\\

Another popular method for training a deep learning model in a differentially private manner is PATE (Private aggregation of teacher ensembles) \cite{papernot_semi-supervised_2017}. PATE is a semi-supervised approach that needs public data. However, in contrast to DP-SGD, PATE can be applied to every machine learning technique. First, several teacher models are trained on separate private datasets. Next, the public data is labeled based on the differentially private aggregation of the predictions of the teacher models. Finally, a student model is trained on the noisily labeled public data, which can be made public afterwards.

An alternative to rendering the deep learning model itself differentially private is output perturbation, where noise is added to the model's predictions. However, with this method, the privacy budget increases with every prediction made making it necessary to limit the number of inferences that can be made with the model. For an example of output perturbation in deep learning see Ye et al. \cite{ye_one_2022}.

\section{Auditing and Evaluation}
\label{sec:auditing}
Auditing and evaluating deep learning models is important to ensure that they provide effective privacy preservation while maintaining utility. As differential privacy is always a trade-off between privacy and utility, privacy evaluation helps choosing a suitable privacy budget: high enough to protect sensitive information but low enough to provide sufficient accuracy. Which level of accuracy is sufficient is application-dependent. In some cases, it might also be important to evaluate not only the utility overall but also on different subgroups present in the training data to detect potential unfairness and biases. In this section, we first discuss the attack-based evaluation of empirical privacy risks, and then the evaluation of accuracy on subgroups.

\subsection{Attack-based Evaluation}

While the differential privacy parameter $\epsilon$ provides an upper bound on the privacy loss, attack-based evaluation can give a lower bound. Even though empirical privacy evaluation cannot provide any guarantee, it can be useful to answer questions regarding the practical meaning of different privacy budgets, or how different aspects (e.g., the applied differential privacy notion) influence the actual privacy risk. Moreover, the considerable gap that was repeatedly observed between the lower and upper bound lead to the assumption that the worst-case setting is too pessimistic and actual privacy is much lower. As a consequence, high privacy budgets (i.e., high $\epsilon$) were often chosen in practical applications \cite{desfontaines_list_2021}. While this choice improved the models' utility, the question remained if the gap between lower and upper privacy bounds implies that the DP guarantee is too pessimistic, or if the applied attacks were just weak and future stronger attacks would be able to fully exploit the privacy budget. Additionally, attacks can be used to evaluate whether models preserve privacy in specific settings, e.g., in cases where a class consists of instances by a single individual. The papers reviewed in this section give insights into these matters. Table \ref{tab:attacksuccessrate} provides an overview of the reviewed works, their applied attacks and evaluation metrics. Table \ref{tab:datasets} shows which datasets were used for auditing.

\begin{table} 
\centering
\caption{Summary of attack-based evaluation methods. The table includes the works reviewed in this chapter and states the used attack(s), if they assume black- or white-box access to the target model, and how the target model's vulnerability was measured. MIA is the abbreviation for membership inference attack. The attacker's advantage is a measure of privacy leakage proposed by Yeom et al. \cite{yeom_privacy_2018}. $\epsilon_{LB}$ and $(\epsilon,\delta)_{LB}$ respectively refers to the empirical lower bound of the privacy budget.}
\label{tab:attacksuccessrate}
\makebox[\textwidth]{\begin{tabular}{p{4.8cm} p{2.5cm} p{3.5cm} p{3.6cm}}
\toprule
\textbf{Work}	& \textbf{Attack} & \textbf{Black- or white-box} & \textbf{Evaluation metric(s)}\\
\midrule
Jayaraman and Evans, 2019 \cite{jayaraman_evaluating_2019} & MIA \& model inversion & black- \& white-box & accuracy loss, attacker's advantage \\
Chen et al., 2020 \cite{chen_differential_2020} & MIA & white-box & accuracy\\
Leino and Fredrikson, 2020 \cite{leino_stolen_2020} & MIA & white-box & accuracy, recall, precision, attacker's advantage\\
Jagielski et al., 2020 \cite{jagielski_auditing_2020} & poisoning attack & black-box & $\epsilon_{LB}$ \\
Nasr et al., 2021 \cite{nasr_adversary_2021} & MIA \& poisoning attacks & black- \& white-box & $(\epsilon, \delta)_{LB}$ \\
Park et al., 2019 \cite{park_attack-based_2019} & model inversion &  black-box & success rate, impact of the attack\\
Zhang et al., 2020 \cite{zhang_secret_2020} & model inversion & white-box & accuracy\\
\bottomrule
\end{tabular}}
\end{table}

\begin{table} 
\centering
\caption{Datasets used for auditing DP-DL in the reviewed papers.}
\label{tab:datasets}
\begin{tabular}{lll}
\toprule
\textbf{Dataset}	& \textbf{Data type} & \textbf{Paper}\\
\midrule
Adult Data Set \cite{adultdata} & tabular & \cite{leino_stolen_2020}\\
AT\&T Database of Faces \cite{databaseoffaces} & images & \cite{park_attack-based_2019} \\
Breast Cancer Wisconsin Data Set \cite{breastcancerdata} & tabular & \cite{leino_stolen_2020}\\
CIFAR-10 \cite{cifar10} & images & \cite{jagielski_auditing_2020,yao_exploration_2022,leino_stolen_2020,nasr_adversary_2021}\\
CIFAR-100 \cite{cifar10} & images & \cite{jayaraman_evaluating_2019,leino_stolen_2020}\\
Fashion-MNIST \cite{xiao_fashion_2017} & images & \cite{jagielski_auditing_2020}\\
German Credit Data \cite{creditdata} & tabular & \cite{leino_stolen_2020}\\
Hepatitis Data Set \cite{hepatitisdata} & tabular & \cite{leino_stolen_2020}\\
Labeled Faces in the Wild \cite{LFWdata} & images & \cite{leino_stolen_2020}\\
MNIST \cite{MNISTdata} & images & \cite{leino_stolen_2020,nasr_adversary_2021,zhang_secret_2020}\\
Pima Diabetes Data Set \cite{diabetesdata} & tabular & \cite{leino_stolen_2020}\\
Purchase-100 \cite{purchasedata} & tabular & \cite{jayaraman_evaluating_2019,jagielski_auditing_2020,nasr_adversary_2021}\\
VGGFace2 dataset \cite{vggface2data} & images & \cite{park_attack-based_2019} \\
Yeast genomic dataset \cite{bloom_genetic_2015} & genomic & \cite{chen_differential_2020}\\
\bottomrule
\end{tabular}
\end{table}

Jayaraman and Evans \cite{jayaraman_evaluating_2019} analyzed how different relaxed notions of differential privacy influence the attack success. They use both membership inference and model inversion attacks, and study DP with advanced composition, zero-concentrated DP and Rényi DP. They conclude that relaxed DP definitions go hand in hand with increased attack success, i.e., privacy loss. That means that using modified privacy analysis can narrow the gap between theoretical privacy guarantee and empirical lower bound.

Chen et al. \cite{chen_differential_2020} confirmed the effectiveness of differential privacy also in case of high-dimensional training data using the example of genomic data. They evaluated a differentially private convolutional neural network (CNN) model with and without sparsity by launching a membership inference attack. Even though model sparsity can improve the model's accuracy in the non-private setting (by mitigating overfitting), they showed that in the private setting it has a negative effect (but improves privacy).

While the previous papers used existing attacks to study different settings, the following works propose novel, stronger attacks.

Leino and Fredrikson \cite{leino_stolen_2020} argued that memorization of sensitive information can not only be apparent in the model's predictions but also in how it uses (externally given or internally learned) features. Features that are only predictive for the training dataset but not for the target data distribution can leak information even in well-generalized models. To evaluate the resulting privacy risk, they proposed a new white-box membership inference attack that also leverages the intermediate representations of the target model. As normal shadow model training does not lead to the same interpretation of internal features (even with identical model architecture, hyperparameters and training data), they linearly approximated each layer, launched a separate attack on each layer and trained a meta-model that combines the outputs of the layer-wise attacks. This novel attack was shown to be more effective than previous attacks. Training the target model with DP-SGD in general decreased the model's vulnerability, but high $\epsilon$ (here: $\epsilon=16$) lead to privacy leakage comparable to the non-private setting.

Jagielski et al. \cite{jagielski_auditing_2020} explicitly analyzed the gap between theoretical privacy guarantee (upper bound) and empirical evaluation (lower bound). The gap can be narrowed either by providing tighter privacy analyses (as discussed in context with Jayaraman and Evans \cite{jayaraman_evaluating_2019}), or by developing stronger attacks (e.g., like Leino and Fredrikson \cite{leino_stolen_2020}). Jagielski et al. \cite{jagielski_auditing_2020} followed the latter approach and showed that the former is approaching its limits. Their novel attack is based on data poisoning, where the attacker purposefully perturbs some training samples to change the parameter distribution (e.g., by adding a white patch in the corner of an image). As this approach is made effective by using poisoning samples that induce large gradients, the gradient clipping of DP-SGD degrades the attack. Their proposed adjustment includes perturbing training samples so that parameters are changed in the direction of their lowest variance. Their estimated lower bound of the privacy loss $\epsilon_{LB}$ was only an order of magnitude below the theoretical upper bound. This suggested the end of continuously tighter privacy guarantees by refined analyses that in the past brought improvements by a thousandfold.

Nasr et al. \cite{nasr_adversary_2021} extended this study further by testing how different assumptions about the attacker's capabilities influence the empirical lower bound. The investigated capabilities include: access to the parameters of the final and all intermediate models, and manipulation of inputs and gradients. In contrast to Jagielski et al. \cite{jagielski_auditing_2020}, they considered approximate DP. Their evaluations revealed that the strongest adversary can exploit the full privacy budget determined by theoretic analysis. They confirmed thereby that the possibilities to improve the privacy-utility trade-off via more advanced privacy analyses are exhausted. However, assuming limited capabilities of the attacker may reduce the upper bound further.

The attacks applied in these papers are all considered typical assessments of DP: Membership inference is the most obvious test of DP arising from its definition; the model inversion attack applied by Jayaraman and Evans \cite{jayaraman_evaluating_2019} tries to infer attributes of specific individuals - which is not possible if the whole record of the individual cannot be recovered; and poisoning attacks exploit the fact that DP has to work even with the worst-case dataset. However, model inversion attacks can also mean the inference of class attributes (instead of individual attributes). In some tasks (e.g., face recognition), a class refers to exactly one individual. The following works \cite{park_attack-based_2019,zhang_secret_2020} analyzed whether DP-DL also protects against this kind of model inversion attack.

Park et al. \cite{park_attack-based_2019} evaluated the privacy loss of a face recognition model by reconstructing the training images from the model's predictions, and automatically measuring the attack success based on the performance of an evaluation model. Their results showed that even high privacy budgets (e.g., $\epsilon=8$) can provide protection against this model inversion attack compared to the non-private setting.

Zhang et al. \cite{zhang_secret_2020} proposed a generative model inversion (GMI) attack also in the face recognition setting. They first trained a GAN (Generative Adversarial Network; see detailed explanation in Section \ref{sec:gan}) on public data to generate realistic images, and then reconstructed the sensitive face regions by finding the values that maximize the likelihood. They showed that DP-SGD could not prevent their attack. They also argued that higher predictive power of the target models goes hand in hand with increased vulnerability to the attack.\\
Further discussions about the relevance of this kind of model inversion attack can be found in Section \ref{sec:discussion}.

\subsection{Evaluation of subgroups: bias and fairness}
\label{sec:bias}
The evaluation of accuracy on subgroups is a standard approach to identify biases. In general, models should avoid the unfair treatment of different groups especially those based on legally protected attributes like gender, religion and ethnicity. Bagdasaryan et al. \cite{bagdasaryan_differential_2019} showed that applying DP-SGD leads not only to an overall accuracy loss but underrepresented classes are disparately affected. While a greater imbalance in the training data seems to increase the accuracy gap, Farrand et al. \cite{farrand_neither_2020} showed that it is significant even in cases of a 30-70\% split, and also for loose privacy guarantees. The observed effect is believed to mainly arise from the clipping of the gradients, which penalizes those samples that result in bigger gradients, i.e., outliers.

\section{Improving the privacy-utility trade-off of DP-DL}
\label{sec:privacy-utility}

 The main challenge of DP-DL is that by setting meaningful privacy guarantees, utility often deteriorates strongly. In recent years, many propositions for improved DP-DL (mostly DP-SGD) were made that can increase accuracy at the same privacy. Table \ref{tab:improvedDPlearning} gives an overview of the proposed approaches. An alternative method would be to provide tighter theoretical bounds for the privacy loss without influencing the learning algorithm. An example of such an approach evaluated on deep learning can be found in Ding et la. \cite{ding_private_2022}. However, as we already established in Section \ref{sec:auditing}, this line of research seems to have reached its limit.\\

\begin{table}
\centering
\caption{Summary of approaches for improving DP-DL.}
\label{tab:improvedDPlearning}
\begin{tabular}{ll}
\toprule
\textbf{Approach} & \textbf{Paper}\\
\midrule
Adapting the model architecture & \cite{papernot_tempered_2020}\\
Improving the hyperparameter selection &  \cite{papernot_tempered_2020,liu_private_2019}\\
Applying feature engineering and transfer learning & \cite{tramer_differentially_2021}\\
Mitigating the clipping bias (to improve convergence) & \cite{chen_understanding_2020}\\
Pruning the model & \cite{gondara_training_2021,adamczewski_differential_2023,phong_differentially_2023}\\
Adding heterogeneous noise & \cite{yu_differentially_2019,xiang_differentially-private_2019,xu_adaptive_2020,adesuyi_layer-wise_2019,gong_preserving_2020}\\
\bottomrule
\end{tabular}
\end{table}

\begin{figure}[h]
\centering
\includegraphics[width=0.8\linewidth]{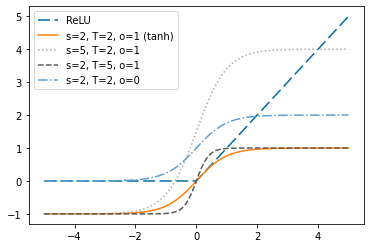}
\caption{Examples of tempered sigmoid functions in comparison with the ReLU function. Tempered sigmoid functions with their parameters $s$, $T$ and $o$ are bounded activation functions proposed by Papernot et al. \cite{papernot_tempered_2020} to improve private deep learning. ReLU is an unbounded activation function commonly used in deep learning.}
\label{fig:activation_funcs}
\end{figure}

One approach to increase the accuracy of DP-SGD at the same privacy is to adapt the architecture of the deep learning model to better suit differentially private learning. Papernot et al. \cite{papernot_tempered_2020} observed that rendering SGD differentially private as proposed by Abadi et al. \cite{abadi_deep_2016} leads to exploding gradients. The larger the gradients, the more information is lost during clipping, which in turn hurts the model's accuracy. To mitigate this effect, Papernot et al. \cite{papernot_tempered_2020} proposed to use bounded activation functions instead of the unbounded ones commonly used in non-private training (e.g., ReLU). They introduced tempered sigmoid activation functions:

\begin{equation}
    \Phi(x) = \frac{s}{1+e^{-T x}}-o
\end{equation}

where the parameter $s$ controls the scale of the activation, the inverse temperature $T$ regulates the gradient norms and $o$ is the offset (see Figure \ref{fig:activation_funcs}). The setting [$s=2$, $T=2$, $o=1$] results in the hyperbolic tangent (tanh) function.

Papernot et al. \cite{papernot_tempered_2020} showed that tempered sigmoids can increase the model's accuracy. For the MNIST and Fashion-MNIST datasets the tanh function performed best.\\

Another important aspect when tuning deep learning models for improved performance is hyperparameter selection (e.g., choosing the learning rate). Even though it might be tempting to transfer the choice of hyperparameters from non-private to private model, Papernot et al. \cite{papernot_tempered_2020} showed that not only the model's architecture but also the hyperparameters should be chosen explicitly for the private model in contrast to using what worked well in the non-private setting. As this can result in additional privacy leakage, one should consider private selection of hyperparameters, as, for example, proposed by Liu et al. \cite{liu_private_2019}.\\

Similar to non-private deep learning, DP-SGD can benefit from feature engineering, additional data, and transfer learning. Tramèr and Boneh \cite{tramer_differentially_2021} showed that handcrafted features can significantly improve the private model's utility compared to end-to-end learning. The comparable increase in accuracy for private end-to-end learning can be achieved by using an order of magnitude more training data or by transferring features learned from public data.\\

While the preceding techniques are already known from non-private deep learning, DP-SGD introduces two new steps that offer opportunities for improvement: gradient clipping and noise addition.

Chen et al. \cite{chen_understanding_2020} found that the bias introduced by gradient clipping can cause convergence issues. They discovered a relationship between the symmetry of the gradient distribution and convergence: Symmetric gradient distributions lead to convergence even if a large fraction of gradients are heavily scaled down. Based on this finding, Chen et al. \cite{chen_understanding_2020} proposed to introduce additional noise before clipping when the gradient distribution is non-symmetric. It is important to note that this approach may lead to better but slower convergence due to the additional variance, and therefore only improves the privacy-utility trade-off in specific use-cases.\\

Another observation specific to DP-SGD is that the privacy-utility trade-off worsens with growing model size. A higher number of model parameters results in a higher gradient norm, meaning that clipping the gradient to the same norm leads to a higher impact. If the clipping norm is also increased, then more noise has to be added to achieve the same privacy guarantee. This effect can be mitigated by either reducing the number of model parameters (parameter pruning) \cite{gondara_training_2021,adamczewski_differential_2023} or compressing the gradients (gradient pruning) \cite{adamczewski_differential_2023,phong_differentially_2023}.

The work by Gondara et al. \cite{gondara_training_2021} is based on the lottery ticket hypothesis  \cite{frankle_lottery_2019,frankle_linear_2020}, which says that there exist sub-networks in large neural networks that when trained separately achieve comparable accuracy as the full network. The term "lottery ticket" refers to the pruned networks and comes from the idea that finding a well-performing sub-network is like winning the lottery. Gondara et al. \cite{gondara_training_2021} altered the original lottery ticket hypothesis to comply with differential privacy. First, the lottery tickets are created non-privately using a public dataset. Next, the accuracy of each lottery ticket is evaluated on a private validation set, and the best sub-network is selected while preserving DP via the Exponential Mechanism. Finally, the winning ticket is trained using DP-SGD.

Adamczewski and Park \cite{adamczewski_differential_2023} proposed DP-SSGD (differentially private sparse stochastic gradient descent) that too relies on model pruning. They experimented with both parameter freezing, where just a subset of parameters are trained, and parameter selection, where a different subset of parameters are updated each iteration. The updated parameters were either chosen randomly or based on their magnitude.

Both of these model pruning approaches rely on publicly available data that should be as similar as possible to the private data. In contrast, Phong and Phuong \cite{phong_differentially_2023} proposed a gradient pruning method that works without public data. Additionally to making the gradients sparse, they use memorization to maintain the direction of the gradient descent.\\

A further line of research deals with adapting the noise that is added to the differentially private model during training. While the original DP-SGD algorithm adds the same amount of noise to each gradient coordinate independent of the training progress, this line of work adds noise either based on the learning progress (e.g., number of executed epochs) \cite{yu_differentially_2019} or based on the coordinates' impact on the model \cite{xiang_differentially-private_2019,xu_adaptive_2020,adesuyi_layer-wise_2019,gong_preserving_2020}.

Yu et al. \cite{yu_differentially_2019} argues that with training progress and therefore convergence to the local optimum the model profits more from smaller noise. This "dynamic privacy budget allocation" is similar to the idea behind adaptive learning rates, which is a common technique in non-private learning and can also be applied in private learning (see for example \cite{xu_adaptive_2020,xu_ganobfuscator_2019}). Yu et al. compared different variants of dynamic schemes to allocate privacy budgets, including predefined decay schedules like exponential decay, or noise scaling based on the validation accuracy on a public dataset. They showed that all dynamic schemes outperform the uniform noise allocation to a similar extent.

Xiang et al. \cite{xiang_differentially-private_2019} treated the privacy-utility trade-off as an optimization problem, namely minimizing the accuracy loss while satisfying the privacy constraints. Consequently, less noise is added to those gradient coordinates that have a high impact on the model's output. While the model's utility was improved for a range of privacy budgets and model architectures, the method is computationally expensive due to the high dimensionality of the optimization problem.

Xu et al. \cite{xu_adaptive_2020} advanced the approach further not only reducing the computational demand but also improving convergence (and therefore decreasing the privacy budget). The improved version called AdaDP (\textbf{ada}ptive and fast convergent approach to \textbf{d}ifferentially \textbf{p}rivate deep learning) replaces the computationally expensive optimization with a heuristic approach to compute the gradient coordinates' impact on the model's output. The added noise is not only adaptive with regards to the coordinates' sensitivity but also decreases with the number of training iterations. Faster convergence is achieved by incorporating an adaptive learning rate that is larger for less frequently updated coordinates.

A related research direction is the usage of explainable AI methods to calibrate the noise. Both Gong et al. \cite{gong_preserving_2020} and Adesuyi et al. \cite{adesuyi_layer-wise_2019} use layer-wise relevance propagation (LRP) \cite{montavon_layer-wise_2019} to determine the importance of the different parameters. Gong et al. \cite{gong_preserving_2020} proposed the ADPPL (adaptive differential privacy preserving learning) framework that adds adaptive Laplace noise \cite{phan_adaptive_2017} to the gradient coordinates according to their relevance. In contrast, the approach by Adesuyi et al. \cite{adesuyi_layer-wise_2019} is based on loss function perturbation. As deep learning models have non-convex loss functions, the polynomial approximation of the loss function is computed before adding Laplace noise. LRP is used to classify the parameters either as high or low relevance, adding small and large noise accordingly.

Explainable AI-based approaches can also be applied in the local DP setting, for example Wang et al. \cite{wang_dnn-dp_2020} uses feature importance to decide how much noise to add to the training data.\\ 

A list of examples for results that the discussed works reported can be found in Table \ref{tab:utility}. For comparison, the results for the original DP-SGD by Abadi et al. \cite{abadi_deep_2016} were included as well. The different accuracies for the three different reported privacy levels from the original DP-SGD paper clearly show the privacy-utility trade-off typical for differentially private algorithms. Direct comparison between the methods is difficult due to different network architectures and hyperparameters, the different evaluation datasets, and the different privacy levels measured according to different DP notions (e.g., $\epsilon$-DP, $(\epsilon,\delta)$-DP, $\rho$-zCDP (zero-concentrated DP)).

\begin{table}
\caption{Selected examples of results for the improved DP methods in comparison with the original DP-SGD by Abadi et al. \cite{abadi_deep_2016}. The used deep learning models include fully-connected neural networks (FCNN) and convolutional neural networks (CNN). Some include a principal component analysis (PCA) layer in front. We reported the model, evaluation dataset, accuracy and privacy budget for all methods that state exact values, preferably on the MNIST dataset.\\
*This result was achieved by parameter freezing. Parameter selection performed slightly worse.\\ 
**This result was achieved by a polynomial decay schedule. Other budget allocation schemes performed comparably.\\ 
***$\rho$ is the privacy parameter for zero-concentrated DP.}
\label{tab:utility}
\makebox[\textwidth]{\begin{tabular}{p{3.6cm} p{3.5cm} p{2.4cm} p{1.5cm} p{2.7cm}}
\toprule
\textbf{Method}	& \textbf{Model} & \textbf{Dataset} & \textbf{Accuracy} & \textbf{Privacy}\\
\midrule
Original \cite{abadi_deep_2016} & PCA + FCNN (1 hidden layer) & MNIST & 90\% & $\epsilon=0.5, \delta=10^{-5}$\\
Original \cite{abadi_deep_2016} & PCA + FCNN (1 hidden layer) & MNIST & 95\% & $\epsilon=2, \delta=10^{-5}$\\
Original \cite{abadi_deep_2016} & PCA + FCNN (1 hidden layer) & MNIST & 97\% & $\epsilon=8, \delta=10^{-5}$\\
\midrule
Tanh \cite{papernot_tempered_2020} & CNN (2 convolutional layers) & MNIST & $98.1\%$ & $\epsilon=2.93, \delta=10^{-5}$ \\
DPLTH \cite{gondara_training_2021} & FCNN (3 hidden layers) & public: MNIST; private: Fashion-MNIST & $76\%$ & $\epsilon=0.4$ \\
DP-SSGD \cite{adamczewski_differential_2023} & CNN (2 convolutional layers) & MNIST & $97.02\%$* & $\epsilon=2$\\
Gradient compression \cite{phong_differentially_2023} & CNN (2 convolutional layers) & MNIST & $98.52\%$ & $\epsilon=1.71, \delta=10^{-5}$ \\
Dynamic privacy budget allocation** \cite{yu_differentially_2019} & PCA + FCNN (1 hidden layer) & MNIST & $93.2\%$** & $\rho=0.78$***\\
Adaptive noise \cite{xiang_differentially-private_2019} & CNN (2 convolutional layers) & MNIST & $94.69\%$ & $\epsilon=1,\delta=10^{-5}$\\
AdaDP \cite{xu_adaptive_2020}& PCA + FCNN (1 hidden layer) & MNIST & $96\%$ & $\epsilon=1.4,\delta=10^{-4}$\\
Noise acc. to LRP \cite{adesuyi_layer-wise_2019}  & FCNN (3 hidden layers) & Winsconsin Diagnosis Breast Cancer (WDBC) & $94\%$ & $\epsilon=1.1$ \\
ADPPL framework \cite{gong_preserving_2020} & CNN (2 convolutional layers) & MNIST & $94\%$ & $\epsilon=1$ \\
\bottomrule
\end{tabular}}
\end{table}

\section{Beyond membership and attribute inference: Applying DP-DL to protect against other threats}
\label{sec:otherthreats}

Differential privacy is usually applied to protect against re-identification. In the context of deep learning this mainly includes membership inference and model inversion attacks. This section reviews cases in which differential privacy can protect against other threats deep learning models exhibit. Table \ref{tab:otherthreats} lists the discussed threats and literature.\\

\begin{table} 
\centering
\caption{Summary of threats other than membership and attribute inference against which DP-DL was applied.\\ *These two approaches are both based on threat identification via anomaly detection.} 
\label{tab:otherthreats}
\begin{tabular}{ll}
\toprule
\textbf{Threat} & \textbf{Paper}\\
\midrule
Model extraction attacks & \cite{zheng_bdpl_2019,zheng_protecting_2022,yan_monitoring-based_2022}\\
Adversarial attacks & \cite{lecuyer_certified_2019,phan_heterogeneous_2019}\\
Privacy risk of interpretable DL & \cite{harder_interpretable_2020}\\
Privacy risk of machine unlearning   & \cite{chen_when_2021}\\
Backdoor poisoning attacks* & \cite{du_robust_2019}\\
Network intrusion* & \cite{yang_griffin_2022}\\
\bottomrule
\end{tabular}
\end{table}

One threat against which differential privacy can be applied even though it was not originally intended for that purpose are model extraction attacks. Model extraction is primarily a security and confidentiality issue but it can also compromise privacy as it facilitates membership and attribute inference.

Zheng et al. \cite{zheng_bdpl_2019,zheng_protecting_2022} observed that most model extraction attacks infer the decision boundary of the target model via nearby inputs. They introduced the notion of boundary differential privacy ($\epsilon$-BDP) and proposed to append a BDP layer to the machine learning model that 1) determines which outputs are close to the decision boundary (a question not straightforward for deep learning models as the decision boundary has no closed form) and 2) adds noise to them. Previous input-output pairs are cached to ensure that the same input results in the same noisy output. This method guarantees that the attacker cannot learn the decision boundary with more than a predetermined level of precision (controlled by $\epsilon$).

A subsequent study by Yan et al. \cite{yan_monitoring-based_2022} showed that without caching the BDP layer cannot protect against their novel model extraction attack. They proposed an alternative to caching: monitoring the privacy leakage and adapting the privacy budget accordingly.\\

Differential privacy can also be adapted to protect deep learning models against adversarial attacks. While originally DP is defined on a record level, feature-level DP can achieve robustness against adversarial examples. For example, Lecuyer et al. \cite{lecuyer_certified_2019} proposed PixelDP, a pixel-level DP layer that can be added to any type of deep learning model. As this approach only guarantees robustness but not differential privacy in the original sense, Phan et al. \cite{phan_heterogeneous_2019} developed the method further by combining it with DP-SGD. To deal with the trade-off between robustness, privacy and utility, they relied on heterogeneous noise: More noise is added to more vulnerable coordinates.\\

Another line of research is the relationship between interpretability/explainability and privacy. Even though interpretable/explainable AI methods are not a threat scenario on their own, they can facilitate privacy attacks. Harder et al. \cite{harder_interpretable_2020} looked at how models can both be interpretable and guarantee differential privacy. Models trained with DP-SGD are not vulnerable to gradient-based interpretable AI methods due to the post-processing property of DP. However, gradient-based methods can only provide local explanations, i.e., about how relevant a specific input is for the model's decision, but Harder et al. \cite{harder_interpretable_2020} was specifically interested in methods that can also give global explanations, i.e., about how the model works overall. They introduced differentially private locally linear maps (DP-LLM), which can approximate deep learning models and are inherently interpretable.\\
 
Chen et al. \cite{chen_when_2021} investigated the privacy risk of machine unlearning. Machine unlearning \cite{cao_towards_2015,villaronga_humans_2018} is the process of removing the impact one or more datapoints have on a trained model. Common methods are retraining from scratch and SISA (Sharded, Isolated, Sliced, and Aggregated) \cite{bourtoule_machine_2021}, where the original model consists of $k$ models each trained on a subset of the training set and therefore only one submodel is retrained for unlearning. While the main idea is to be able to comply with privacy regulations like the right to be forgotten in the European General Data Protection Regulation (GDPR), the unlearning can disclose additional information about the removed datapoint(s). Chen et al. \cite{chen_when_2021} proposed a new black-box membership inference attack that exploits both the original and the unlearned model. Their attack was more powerful than classical membership inference attacks, and also worked for well-generalized models, when several datapoints were removed, when the attacker missed several intermediate unlearned models, and when the model was updated with new inputs. They showed that DP-SGD is an effective defense against the privacy risk of machine unlearning.\\

Instead of protecting the model directly, differential privacy can also be used to improve the detection of attacks \cite{du_robust_2019,yang_griffin_2022}. This approach can be viewed as a kind of anomaly detection, where the attack scenario is the outlier/novelty. Anomaly detection with deep learning models (e.g., autoencoders, CNNs) is based on the model's tendency to underfit on underrepresented subgroups, i.e., the model's error is expected to be higher for atypical inputs. Training the model with differential privacy amplifies this effect (see Section \ref{sec:bias}). While this leads to negative consequences in the context of fairness and bias, here it can be used to improve the performance of anomaly detection.

Du et al. \cite{du_robust_2019} applied this approach on crowdsourcing data, that is, data stemming from many individuals. These individuals could launch a backdoor poisoning attack by maliciously adapting their contributed samples. To protect the target model, the poisoned samples need to be identified and removed from the training set. They showed that differential privacy can improve the performance of anomaly detection in this context. Based on the same reasoning, Yang et al. \cite{yang_griffin_2022} proposed Griffin, a network intrusion detection system.

\section{Differentially private generative models}
\label{sec:gan}
Generative models are a class of machine learning models that aim to generate new data samples similar to the data samples from the training set. The synthetic data created with generative models are often seen as privacy-preserving as they are not directly linked to real entities or individuals. However, similarly to other machine learning models, generative models can memorize sensitive information and be vulnerable to privacy attacks. This section gives an overview of recent works regarding the generation of differentially private synthetic data with generative models (see Table \ref{tab:generativemodels}).

\begin{table} 
\centering
\caption{Summary of differentially private generative models. The methods are either based on (variational) autoencoders or Generative Adversarial Networks (GAN). The DP algorithm DP-EM refers to differentially private expectation maximization \cite{park_dp-em_2017}.}
\label{tab:generativemodels}
\begin{tabular}{lll}
\toprule
\textbf{Method} & \textbf{Type of generative model(s)    } & \textbf{DP algorithm} \\
\midrule
DPGM \cite{acs_differentially_2019} & variational autoencoders & DP k-means \& DP-SGD\\
DP-SYN \cite{abay_privacy_2019} & autoencoders & DP-SGD \& DP-EM\\
PPGAN \cite{liu_ppgan_2019} & GAN & DP-SGD\\
GANobfuscator \cite{xu_ganobfuscator_2019}    & GAN & DP-SGD\\
GS-WGAN \cite{chen_gs-wgan_2020} & GAN & DP-SGD\\
RDP-CGAN \cite{torfi_differentially_2022} & GAN & DP-SGD\\
PATE-GAN \cite{jordon_pate-gan_2019} & GAN & PATE\\
\bottomrule
\end{tabular}
\end{table}

\subsection{Autoencoders}
An autoencoder is a deep learning model consisting of two parts: an encoder and a decoder. The encoder learns to map the input to a lower-dimensional space, while the decoder learns to reconstruct the input from this intermediate representation. Variational autoencoders (VAEs) work according to the same principle but learn the probability distribution of the intermediate (encoded) representations.\\

Acs et al. \cite{acs_differentially_2019} proposed DPGM (Differentially Private Generative Model) based on a mixture of variational autoencoders (VAEs). First, the training data are privately clustered using differentially private version of k-means. Next, one VAE per cluster is trained with DP-SGD. The data distributions learned by the VAEs are used to generate synthetic data. Splitting the input data into clusters and learning separate models has two advantages: Firstly, the models learn faster as they are trained on similar datapoints so less noise is added and the models achieve higher accuracy. Secondly, unrealistic combinations of clusters are avoided.

Abay et al. \cite{abay_privacy_2019} used a similar approach also based on partitioning the training data. In contrast to DPGM \cite{acs_differentially_2019}, they assume a supervised setting and split the data according to their labels. Each class is then used to train a separate autoencoder via DP-SGD. Synthetic data is generated by sampling the intermediate representations with differentially private expectation maximization (DP-EM) \cite{park_dp-em_2017}, and applying the decoder on the new representations. This method (called DP-SYN) outperformed DPGM for most tested datasets.

\subsection{GANs}
Generative Adversarial Networks (GANs) consists of two neural networks that play an adversarial game: The generator tries to create synthetic samples similar to the training data, while the discriminator attempts to distinguish between the artificially generated and the real samples. After training, the generator can be used to create synthetic data.\\

In the original GAN, the generator is trained based on the Jensen-Shannon distance as a distance measure between the two data distributions. This can lead to instability issues, in particular vanishing gradients, where the generator learns too slowly compared to the discriminator. To mitigate this problem, the Wasserstein GAN (WGAN) \cite{arjovsky_wasserstein_2017} was introduced, which relies on the Wasserstein distance instead of the Jensen-Shannon distance. For this variant, the weights are clipped after each update to make the Wasserstein distance applicable in this context. 
A differentially private version of WGAN (called DPGAN) was first introduced in 2018 by Xie et al. \cite{xie_differentially_2018} exploiting the fact that WGAN already has bounded gradients. Therefore, differential privacy can be achieved by adding noise to the discriminator's gradients without the need to clip them. As the generator has only access to information about the training data via the (now private) discriminator, it is not necessary to train the generator with DP-SGD. Like DPGAN, the following works \cite{liu_ppgan_2019,xu_ganobfuscator_2019,chen_gs-wgan_2020,torfi_differentially_2022} are private variants of WGAN.

The PPGAN by Liu et al. \cite{liu_ppgan_2019} is similar to DPGAN but uses a different optimization algorithm (mini-batch SGD instead of RMSprop). Xu et al. \cite{xu_ganobfuscator_2019} and Chen et al. \cite{chen_gs-wgan_2020} both proposed differentially private GANs based on the improved WGAN by Gulrajani et al. \cite{gulrajani_improved_2017}. This version of WGAN adds a penalty on the gradient norm to the objective function instead of clipping the weights to improve the GANs stability. While the GANobfuscator by Xu et al. \cite{xu_ganobfuscator_2019} trains the whole discriminator with DP-SGD, the GS-WGAN by Chen et al. \cite{chen_gs-wgan_2020} only sanitizes the gradients that propagate information from the discriminator back to the generator, resulting in a more selective gradient perturbation. Moreover, the GANobfuscator includes an adaptive clipping norm (based on average gradient norm on public data) and an adaptive learning rate (based on gradients' magnitudes).

Torfi et al. \cite{torfi_differentially_2022} introduced RDP-CGAN (Rényi DP and Convolutional GAN) addressing two challenges that can make generating synthetic data difficult: 1) mixed discrete-continuous data, and 2) correlated data (temporal correlations or correlated features). These properties are specially prevalent in health data. RDP-CGAN handles discrete data by adding an unsupervised feature learning step in the form of a convolutional autoencoder to the WGAN. This autoencoder is trained using DP-SGD to map the (discrete) input space to a continuous space. Correlations in the data are considered using one-dimensional convolutional layers.\\

Jordon et al. \cite{jordon_pate-gan_2019} took another approach and proposed a differentially private GAN based on the PATE framework. $k$ teacher discriminators are trained to distinguish real and synthetic data samples. The teachers are then used to label the training data for the student discriminator (by noisy aggregation of their predictions). A main contribution is that PATE-GAN sidesteps the need for public data of the original PATE method. This is essential in this setting as the generation of synthetic data is usually necessary exactly because no public data is available. They could show that their approach outperforms DPGAN.

\section{Specific applications of DP-DL}
\label{sec:applications}

Differentially private deep learning (DP-DL) has attracted increased interest in a wide range of application areas. Table \ref{tab:specificapplications} lists the application fields and corresponding papers discussed in this survey.

\begin{table} 
    \centering
    \caption{Summary of application areas of DP-DL.}
    \label{tab:specificapplications}
    \begin{tabular}{ll}
        \toprule
        \textbf{Application} & \textbf{Paper}\\
        \midrule
        Image publishing & \cite{yu_gan-based_2021,wen_identitydp_2022}\\
        Medical image analysis & \cite{muftuoglu_differential_2020,wu_p3sgd_2019}\\
        Face recognition & \cite{li_privacy-preserving_2019,chamikara_privacy_2020}\\
        Video analysis & \cite{cangialosi_privid_2022,giorgi_privacy-preserving_2022}\\
        Natural language processing & \cite{li_large_2022,lyu_towards_2020,fernandes_generalised_2019,feyisetan_leveraging_2019,ahmed_prch_2020}\\
        Smart grid & \cite{ustundag_soykan_differentially_2019,abdalzaher_data_2022}\\
        Recommender systems & \cite{zhang_graph_2021,chen_differential_2022}\\
        Mobile devices & \cite{wang_private_2019,li_privacy-preserving_2019,chen_rnn-dp_2020,jiang_utility-aware_2021}\\
        \bottomrule
    \end{tabular}
\end{table}

\subsection{Image publishing}
Nowadays, a high amount of images is getting published on a daily basis via the internet. In particularly privacy-sensitive cases, de-identification techniques like blurring and pixelation are used to protect certain objects (e.g., faces and license plates). However, these methods compromise the images' quality and usability. Yu et al. \cite{yu_gan-based_2021} and Wen et al. \cite{wen_identitydp_2022} proposed to replace the sensitive image content with synthetic data in a differentially private manner. They both use GANs and introduce Laplace noise into the latent representations. While Yu et al. \cite{yu_gan-based_2021} first applied a CNN to detect the sensitive image regions, Wen et al. \cite{wen_identitydp_2022} focused specifically on face anonymization making this step redundant. Instead, they concentrated on preserving the visual traits by encoding the attribute and identity information separately and only perturbing the latter.

\subsection{Medical image analysis}
Privacy is of particular relevance in the health domain. One application area where DP was recently applied is medical image analysis, e.g., for COVID-19 diagnoses from chest X-rays \cite{muftuoglu_differential_2020} and for classification of histology images \cite{wu_p3sgd_2019}. The former applied PATE on a convolutional deep neural network. The latter introduced P3SGD (patient privacy preserving SGD), a variant of DP-SGD that protects patient-level instead of image-level privacy. They showed that differential privacy can not only preserve privacy but also mitigate overfitting in cases of small numbers of training records.

\subsection{Face recognition}
\label{sec:face_rec}
Face recognition is a prevalent technology for security, e.g., for unlocking smartphones and for surveillance. For face recognition algorithms, not only privacy but also performance is critical as they are often deployed on devices with limited resources and the analyses should be carried out in real-time. To this end, methods that decrease the size of the models while preserving differential privacy were proposed.

Li et al. \cite{li_privacy-preserving_2019} introduced LightFace, a lightweight deep learning model designed for private face recognition on mobile devices. The approach uses depth-wise separable convolutions for model size reduction, and a Bayesian GAN and ensemble learning for privacy preservation. They were able to decrease the model size and the computational demand while still outperforming both DP-SGD and PATE. Chamikara et al. \cite{chamikara_privacy_2020} proposed PEEP (\textbf{p}rivacy using \textbf{e}ig\textbf{e}nface \textbf{p}erturbation), where the dimensionality of the images for training the deep learning model is reduced with differentially private PCA (Principal Component Analysis).

\subsection{Video analysis}
With the prevalence of CCTV cameras, automatic video analyses are on the rise, including traffic monitoring and security surveillance. Compared to image analysis, video analysis is more complex due to the additional time dimension. Private traffic monitoring tries to answer questions like how many people or cars were passing in a certain time period or how long people or cars were visible on average, while disclosing no information about individuals (e.g., if a specific person or car was observed). To this end, Cangialosi et al. \cite{cangialosi_privid_2022} introduced the differentially private video analytics system Privid. Privacy-preserving video analysis is also relevant in security surveillance, e.g., for crime and threat detection. Giorgi et al. \cite{giorgi_privacy-preserving_2022} proposed training an autoencoder with DP-SGD for detecting anomalies (e.g., vandalism, robbery, assault) in CCTV footage.

\subsection{Natural language processing}
Differential privacy is also increasingly applied in Natural language processing (NLP), protecting against different threat scenarios like membership inference of a phrase or word, author attribute inference (e.g., age or gender), authorship identification, or disclosure of sensitive content. Depending on the use-case and threat scenario, the unit of privacy can be, e.g., a token, a word, a sentence, a document, or a user. Differentially private NLP models can be achieved in two ways: 1) by training the model with DP-SGD \cite{li_large_2022}, or 2) by perturbing the text representations further used for training \cite{lyu_towards_2020,fernandes_generalised_2019,feyisetan_leveraging_2019}.

Applying the original DP-SGD on large language models can lead to deficient accuracy and unreasonably high computational and memory overhead. Pretraining, refined hyperparameter selection and finetuning can improve the performance significantly \cite{li_large_2022}. Moreover, Li et al. \cite{li_large_2022} proposed "ghost clipping", which avoids computing the per-example gradients explicitly and infers the per-example gradient norms in a less memory-demanding way.

The perturbation of text representations is a local DP method. Even though it can be used in centralized deep learning, it is primarily applied in the distributed setting where the server training the model is untrusted. Additionally to pure text processing, text representation perturbation can also be applied to related tasks, e.g., speech transcription \cite{ahmed_prch_2020}.

\subsection{Smart energy networks}
Smart energy networks, also called smart grids, use digital technologies to optimize the generation, distribution and usage of electricity. Key components are smart meters, which allow real-time monitoring of energy consumption and, thereby, load forecasting. However, smart meter data can disclose personal information, e.g., daily habits of residents.

Abdalzaher et al. \cite{abdalzaher_data_2022} provided an overview of privacy-related issues of smart meters and possible defense mechanisms including differential privacy. While they focused on private data release, Ustundag Soykan et al. \cite{ustundag_soykan_differentially_2019} proposed a private load forecasting method on smart meter data using DP-SGD.

\subsection{Recommender systems}
Recommender systems are programs that provide personalized recommendations to users, based on their past behaviors and contextual information (so called user features). Similar to other machine learning models, recommender systems are at risk of leaking personal information.

Zhang et al. \cite{zhang_graph_2021} proposed a recommender system based on a graph convolutional network that protects both the user-item interactions (modelled as a graph) and the additionally used user features. The former is achieved by perturbation of the model's (polynomially approximated) loss function; the latter by adding noise directly to the user features.

Chen et al. \cite{chen_differential_2022} focused on cross-domain recommendation, where knowledge is transferred from one domain to another, usually because the target domain lacks enough data. The information transfer introduces a privacy risk for the users of the source domain. Chen et al. proposed a solution based on differentially private rating publishing and subsequent recommendation modeling with deep neural networks.

\subsection{Mobile devices}
The combination of differential privacy and deep learning was also studied in the context of mobile devices. On the one hand, studies looked at how to reduce the model's size and computational demand to allow the deployment and/or training of deep neural networks on devices with limited resources while still preserving utility and training privacy. For example, Wang et al. \cite{wang_private_2019} proposed an architecture-independent approach based on hint learning and knowledge distillation, and Li et al. \cite{li_privacy-preserving_2019} introduced the lightweight face recognition algorithm LightFace already discussed in Section \ref{sec:face_rec}. On the other hand, differential privacy was applied on location-based services - a typical application class for mobile devices \cite{chen_rnn-dp_2020,jiang_utility-aware_2021}.

\section{Discussion and future directions}
\label{sec:discussion}

This study reviews the latest developments on differential privacy in centralized deep learning. The main research focuses of the last years were: 1) auditing and evaluation, 2) improvements of the privacy-utility trade-off, 3) applications of DP-DL to protect against threats other than membership and attribute inference, 4) differentially private generative models, and 5) specific application domains. For each subtopic, we provided a comprehensive summary of recent advances. In this last section, we discuss the key points, interconnections and expected future directions of the respective topics and differentially private centralized deep learning in general.

Auditing and evaluating deep learning models is a key research topic not only but especially for differentially private models. We expect the trend of novel attacks to continue, analyzing new threat scenarios and improving our understanding of which aspects influence the attack success. An important element of evaluation is the used dataset. Interestingly, many commonly used datasets for auditing DP-DL (e.g., MNIST, CIFAR-10 or CIFAR-100; see Table \ref{tab:datasets}) are not "relevant to the privacy problem" \cite{cummings_challenges_2023}. There is a need for more realistic benchmark datasets that include private features. Another point to consider is that with the rise of continual learning, auditing is not only relevant once before deployment but should be carried out repeatedly whenever the training set changes \cite{cummings_challenges_2023}. This is also the case when machine unlearning is applied (as discussed in Section \ref{sec:otherthreats}).

The accuracy disparity between subgroups is one of many biases that are studied in the field of fair AI - with the goal of avoiding discriminatory behavior of machine learning models. Its amplification by differential privacy, the underlying causes and possible mitigation strategies are actively researched. For example, de Oliviera et al. \cite{de_oliveira_empirical_2023} suggested that they are preventable by better hyperparameter selection.

Additionally to empirical assessment, theoretical privacy analysis might be able to provide more realistic upper bounds by including additional assumptions (e.g., about the attacker's capabilities) or features (e.g., the clipping norm or initial randomness \cite{jagielski_auditing_2020}). This could also improve the privacy-utility trade-off.

We also anticipate further research on novel strategies or advancement on existing approaches to improve the privacy-utility trade-off. Some of the mentioned methods could be combined in the future, for example, the differentially private lottery ticket hypothesis approach by Gondara et al. \cite{gondara_training_2021} can be combined with tempered sigmoid activation functions \cite{papernot_tempered_2020}. Additional effort should be made to compare different methods to identify the best performing method(s). Simply summarizing the reported results, as we did in Table \ref{tab:utility}, can not provide sufficient insight. Fair comparison would require testing the methods on the same model (i.e., same architecture and hyperparameters) with the same evaluation dataset for the same privacy level.

When comparing the different approaches, it is also important to note that some rely on public data \cite{tramer_differentially_2021,gondara_training_2021,adamczewski_differential_2023,yu_differentially_2019}. On the one hand, public data might not be available and, therefore, prevent the application of those methods. On the other hand, it is debatable whether public availability justifies disregarding all privacy considerations (see \cite{cummings_challenges_2023,tramer_considerations_2022} for further information).

Section \ref{sec:otherthreats} showed that the concept of differential privacy can be beneficial in diverse threat scenarios. Especially the connection to robustness and explainability might gain importance through the growing interest in trustworthy AI.

The increasing awareness of privacy concerns in combination with the many open questions regarding ethical, legal and methodical aspects make using synthetic data a tempting alternative to applying privacy-enhancing technologies to private data. However, it is important to spread the knowledge that synthetic data alone is not by default privacy-preserving \cite{stadler_synthetic_2022}. Additional protection might be necessary. Even though differentially private synthetic data can be a viable solution, future research is needed, for example, to find good ways to evaluate the usefulness of the data.

This survey demonstrated how diverse the methods and applications of differential privacy can be on the example of deep learning models. While this flexibility is one of the strengths of differential privacy, it can also be a hindrance to its broad deployment due to insufficient understanding. The efforts to make DP more accessible to a wider audience and to promote its (correct) application should continue. That includes not only discussions about how to choose the method, the unit of privacy, and the privacy budget, but also how to verify that implementations are correct (see \cite{kifer_guidelines_2020} and references therein for more information). An example where an implementation error lead to privacy issues even though theoretically DP was proven can be found in \cite{tramer_debugging_2022}.

Additional to the properties inherent to differential privacy that make it hard to understand for laypeople (see Section \ref{sec:DP}), names that are used ambiguously in the research field can add to the confusion. For example, model inversion can refer to inferring 1) the attribute of a single record, or 2) the attribute of a class. While the first obviously implies a privacy concern, the second is primarily problematic if a class consists of instances of one individual, e.g., as is the case in face recognition. As a result of this ambiguous meaning, contradicting conclusions emerged about whether differential privacy protects against model inversion attacks. Some (e.g., \cite{yeom_privacy_2018}) used the first interpretation and concluded that DP naturally also protects against model inversion. Others (e.g., \cite{zhang_secret_2020}) used the second interpretation and showed that DP does not always do so. Interestingly, Park et al. \cite{park_attack-based_2019} also uses the second interpretation but concluded that DP mitigates model inversion attacks. Maybe this is because DP decreases the accuracy of the target model, and, as Zhang et al. \cite{zhang_secret_2020} argued, predictive power and vulnerability to model inversion go hand in hand. Future research should pay attention to accurately define the used terms, and, ideally, the research community should agree on a coherent taxonomy.\\

With the rise of deep learning used in real-world scenarios, new challenges arise. For example, real-world datasets often contain various dependencies, where domain knowledge is required for their correct interpretation. Causal models \cite{pearl_book_2018} may help to capture and model this domain knowledge and additionally improve interpretability \cite{moraffah_causal_2020} and act as guide to avoid biases \cite{makhlouf_survey_2020}. However, they may have new implications on privacy. Growing interest is not only coming from the research and industrial community, but also the public is actively engaging in discussions about the impact of AI applications on society. Most recently large language models like ChatGPT \cite{chatgpt} are in the spotlight - among other things due to privacy concerns. The future will show whether differential privacy will be part of the next generation of deep learning deployments.

All in all, differentially private deep learning achieved significant progress in recent years but open questions are still numerous. We expect the interest in the topic to increase further, especially as new standards and legal frameworks arise. On the way to trustworthy AI, we need not only technical innovations but also legal and ethical discussions about what privacy preservation means in the digital age.

\section{Conclusion}
This survey provides a comprehensive overview of recent trends and developments of differential privacy in centralized deep learning. Throughout the paper, we highlight the different research focuses of the last years, including auditing and evaluating differentially private models, improving the trade-off between privacy and utility, applying differential privacy methods to threats beyond membership and attribute inference, generating private synthetic data, and applying differentially private deep learning models to different application fields. A total of six insights have been derived from literature: (1) A need for more realistic benchmark datasets with private features. (2) The necessity for repeated auditing. (3) More realistic upper privacy bounds would be possible by including additional attack assumptions and model features. (4) Privacy-utility trade-offs can be improved by better comparison of existing methods and a possible combination of best approaches. (5) By default synthetic data is not privacy-preserving and differentially private synthetic data requires more research. (6) Ambiguously used terms lead to confusion in the research field and a coherent taxonomy is needed.

In summary, we explore the advancements, remaining challenges, and future prospects of integrating mathematical privacy guarantees into deep learning models. By shedding light on the current state of the field and emphasizing its potential, we hope to inspire further research and real-world applications of differentially private deep learning.

\section*{Acknowledgments}
The research leading to these results has received funding from the AI for Green programme (Grant: 4352956).  AI for Green is a research, technology and innovation funding programme of the Republic of Austria, Ministry of Climate Action (BMK). The Austrian Research Promotion Agency (FFG) has been authorised for the programme management. Additionally, this work was supported by the "DDAI" COMET Module within the COMET – Competence Centers for Excellent Technologies Programme, funded by the Austrian Federal Ministry (BMK and BMDW), the Austrian Research Promotion Agency (FFG), the province of Styria (SFG) and partners from industry and academia. The COMET Programme is also managed by FFG.

\end{document}